\documentclass[runningheads]{llncs}
\usepackage{graphicx}
\usepackage[euler]{textgreek}
\usepackage{booktabs}
\usepackage[english]{babel}
\usepackage[utf8]{inputenc}
\usepackage{algorithm}
\usepackage{algpseudocode}
\usepackage{pifont}
\usepackage{amsmath}

\usepackage{array}
\usepackage{multirow}
\begin{document}
\title{Double Attention-based Lightweight Network for Plant Pest Recognition}
\titlerunning{Lightweight Network for Plant Pest Recognition}
%
\author{Sivasubramaniam Janarthan \and
	Selvarajah Thuseethan \and
	Sutharshan Rajasegarar \and
	John Yearwood}
\authorrunning{S. Janarthan et al.}
%
\institute{Deakin University, Geelong, VIC 3220, Australia \\
	\email{\{jsivasubramania,tselvarajah,srajas,john.yearwood\}@deakin.edu.au}\\
}
\maketitle

\begin{abstract}
Timely recognition of plant pests from field images is significant to avoid potential losses of crop yields. Traditional convolutional neural network-based deep learning models demand high computational capability and require large labelled samples for each pest type for training. On the other hand, the existing lightweight network-based approaches suffer in correctly classifying the pests because of common characteristics and high similarity between multiple plant pests. In this work, a novel double attention-based lightweight deep learning architecture is proposed to automatically recognize different plant pests. The lightweight network facilitates faster and small data training while the double attention module increases performance by focusing on the most pertinent information. The proposed approach achieves 96.61\%, 99.08\% and 91.60\% on three variants of two publicly available datasets with 5869, 545 and 500 samples, respectively. Moreover, the comparison results reveal that the proposed approach outperforms existing approaches on both small and large datasets consistently.

\keywords{Plant Pest Recognition \and Double Attention \and Lightweight Network \and Deep Learning}
\end{abstract}

\section{Introduction}
\label{sec:introduction}
Plant pests cause severe damage to crop yields, resulting in heavy losses in food production and to the agriculture industry. In order to reduce the risk caused by plant pests, over the years, agricultural scientists and farmers tried various techniques to diagnose the plant pests at their early stage. Although many sophisticated automatic pest recognition algorithms have been proposed in the past, farmers continue to rely on traditional methods like manual investigation of pests by human experts. This is mainly because of poor classification ability and limited in-field applicability of automatic pest recognition systems \cite{wang2022automatic}. Different plant pests share common characteristics, which makes automatic pest recognition a very challenging task and hence traditional handcrafted feature extraction based approaches often failed to correctly classify pests \cite{fina2013automatic}. While conventional deep learning-based techniques achieved benchmark performances in pest classification, they have limited usage with resource constraint devices due to their high computational and memory requirements \cite{alvarez2016learning}. Large labelled data requirement for training is another flip side of conventional deep learning techniques \cite{krizhevsky2012imagenet}. Recently proposed semi-supervised learning of deep networks is also not ideal for this problem as they frequently demonstrate low accuracies and produce unstable iteration results.

In order to prevent the deep model from overfitting, it is also paramount to provide sufficient data during the training phase \cite{gidaris2018dynamic}. However, constructing a large labelled data in the agriculture domain, especially for plant pests, requires not only high standard of expertise but also time-consuming. Moreover, inaccurate labelling of the training data produces deep models with reduced reliability. Few-shot learning concept is proposed simply by replicating humans' ability to recognize any objects with the help of only a few examples \cite{lake2011one}. Few-shot learning has gained popularity across various domains as it can address the classification task with a few training samples. In few-shot learning, the classification accuracy increases as the number of shots grow. However, a major limitation of few-shot learning is that the prediction accuracy drops when the number of ways increases \cite{wang2020generalizing}. Directly applying the classification knowledge learned from meta-train classes to meta-test classes is mostly not feasible, which is another fundamental problem of this approach \cite{ye2020few}.

Constructing decent-performing models with the reduced number of trainable parameters by downsizing the kernel size of convolutions (e.g., from $3 \times 3$ to $1 \times 1$ as demonstrated in \cite{iandola2016squeezenet}) is a significant step towards the development of lightweight networks. In recent years, lightweight deep network architectures have gained growing popularity as an alternative to traditional deep networks \cite{zhang2021att,zhou2020lightweight,zhang2020lightweight,rashid2020ripnet}. The MobileNets \cite{howard2017mobilenets} and EfficientNets \cite{tan2019efficientnet} families are thus far two most widely used lightweight networks. Several lightweight deep network-based techniques have also been proposed for real-time pest recognition \cite{yang2021automated,zha2021lightweight}. The lightweight architectures however suffer to reach the expected level of classification accuracy as they are essentially developed for faster and lighter deployment by sacrificing the performance.

\begin{figure*}[t]
	\begin{center}
		\includegraphics[width=\textwidth]{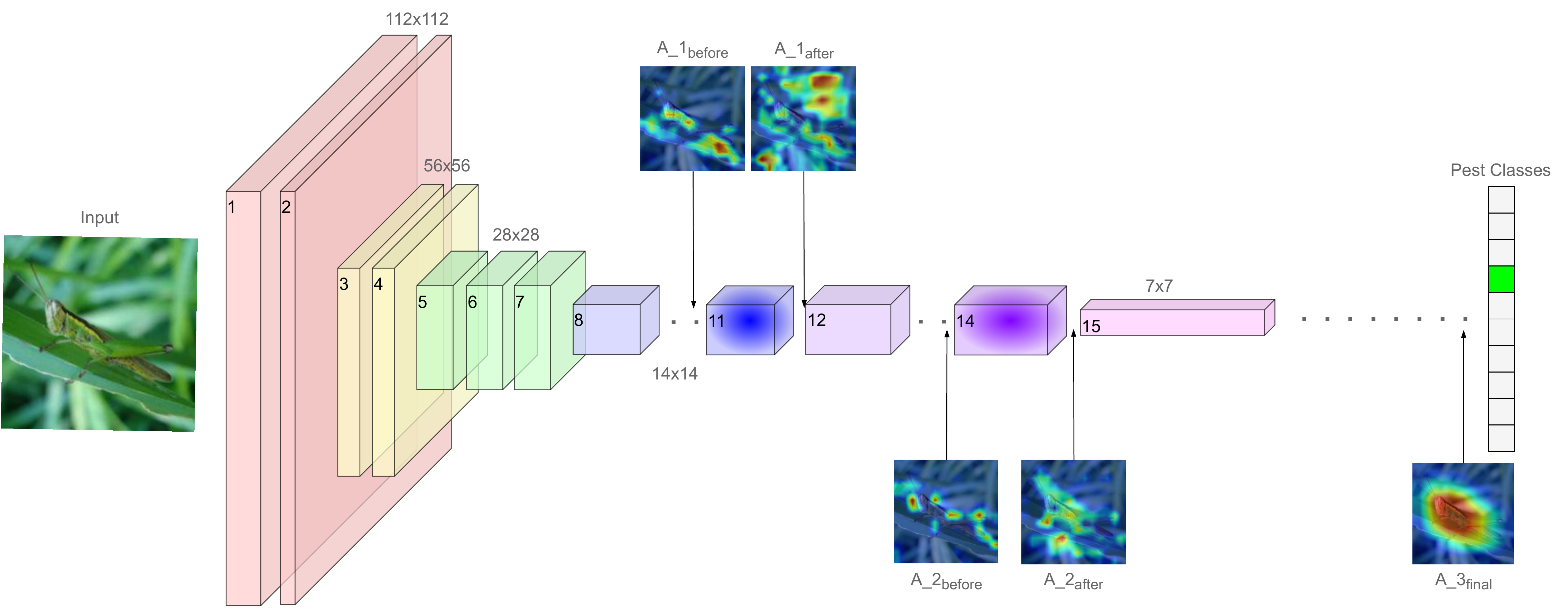}
	\end{center}
	
	\caption{The overall architecture of the proposed double attention-based lightweight pest recognition framework. The layers are labelled to match with the layer numbers indicated in Table \ref{table:proposedapproach}. Some layers of the proposed architecture are avoided in this diagram for brevity. The $A\_1_{before}$, $A\_1_{after}$, $A\_2_{before}$, $A\_2_{after}$ and $A\_3_{final}$ are the activation maps generated before and after respective layers as indicated.}
	
	\label{fig:overallarchitecture}
\end{figure*}

Considering this issue, a novel high-performing and lightweight pest recognition approach is proposed in this study, as illustrated in Figure \ref{fig:overallarchitecture}. While preserving the lightweight characteristic of the deep network, a double attention mechanism is infused to enhance the classification performance. As the attention closely imitates the natural cognition of the human brain, the most influential regions of the pest images are enhanced to learn better feature representations. Notably, attention-aware deep networks have shown improved performances in various classification tasks \cite{wang2017residual,takalkar2021lgattnet}. The key contributions of this paper are three-fold:

\vspace{-0.2cm}

\begin{itemize}
	\item A novel lightweight network-based framework integrated with a double attention scheme is proposed for enhancing the in-field pest recognition, especially using small training data.
	
	\item A set of extensive experiments were conducted under diverse environments to reveal the feasibility and validate the in-field applicability of the proposed framework. To organize diverse environments, three publicly available datasets consisting of small to large number of pest samples are utilized.
	
	\item A comparative analysis is performed to show the superiority of this framework over existing state-of-the-art lightweight networks that are often used for pest recognition methods available in the literature.
\end{itemize}

The remainder of this paper is constructed as follows. The recent advancements of deep learning based pest recognition approaches are given in Section \ref{sec:relatedwork}. Section \ref{sec:experiments} provides comprehensive details on the proposed framework and Section \ref{sec:conclusion} concludes the paper with some future directions.
\vspace{-0.2cm}
\section{Related work}
\label{sec:relatedwork}
Effective pest recognition is essential for preventing the spread of crop diseases and minimizing economic losses in relation to agriculture. Over the years, researchers invested considerable effort to develop pest recognition techniques. Probe sampling, visual inspection and insect trap are some of the widely used manual pest recognition approaches that are still the farmers' favourites when it comes to in-field tasks \cite{banga2018techniques}. Early studies targetted the sounds emitted by the pests to perform the classification, but the paradigm has quickly shifted to digital images in the last decade \cite{liu2017review}. As this paper reviews a few recently proposed prominent pest recognition works, readers are recommended to read \cite{ngugi2021recent} and \cite{nagar2020comprehensive} for a comprehensive pest recognition literature.

With the recent development of deep learning, the automatic pest recognition techniques have become a rapidly growing agricultural research direction \cite{turkouglu2019plant,rustia2021automatic,faisal2021pest}. Different deep convolutional neural networks (CNNs) (i.e., state-of-the-art and tailor-made networks) and capsule networks are prominently used to develop high-performing pest recognition methods. In \cite{chen2021deep}, the pre-trained VGG-16 deep network is exploited to perform tea tree pest recognition. In a similar work, transfer learning is applied on VGG-16 and Inception-ResNet-v2 to enhance the accuracy of the pest identification system \cite{liu2022improved}. An ensemble of six pre-trained state-of-the-art deep networks with majority voting showed improved performance in pest classification \cite{turkoglu2022plantdiseasenet}. Zhang et al. \cite{zhang2022crop} presented a compelling modified capsule network (MCapsNet) for improving the crop pest recognition. A capsule network along with a multi-scale convolution module is adapted to construct novel multi-scale convolution-capsule network (MSCCN) for fine-grained pest recognition \cite{xu2022multi}. Even though these methods showed better pest classification accuracies, they are limited in in-field environments because of their heaviness and large training data requirement. Additionally, the pest recognition methods that use CNNs and capsule networks mostly performed poorly with imbalanced data.

Recently, few-shot and semi-supervised learning mechanisms have been extensively used for plant pest and disease recognition. For instance, Li and Yang \cite{li2020few} introduced a few-shot pest recognition approach where a deep CNN model integrated with the triplet loss is trained and validated. According to the results, the proposed few-shot learning approach showed high generalization capability. In \cite{rustia2021online}, the online semi-supervised learning is exploited for an insect pest monitoring system that achieved a substantial improvement in the accuracy. Despite the fact that the aforementioned methods alleviate large data and class imbalance problems to a certain extent, they continue to show limited classification accuracy.

With the wide adaption of lightweight architectures in image classification tasks \cite{howard2017mobilenets}, there have been many lightweight deep network-based pest recognition proposed in the last two years. It has been confirmed in the literature that the lightweight deep networks can not only be deployed in resource-constrained devices but also be trained with small data \cite{tan2019efficientnet,janarthan2020deep}. In \cite{yang2021automated}, an improved lightweight CNN is introduced by linking low-level and high-level network features that eventually connect rich details and semantic information, respectively. Zha et al. \cite{zha2021lightweight} proposed a lightweight YOLOv4-based architecture with MobileNetv2 as the backbone network for pest detection. In another work, the proposed lightweight CNN model's feasibility in classifying tomato pests with imbalanced data is validated \cite{liang2021imbalance}. The lightweight pest recognition approaches use optimized deep models that mostly compromise the performance.

The attention mechanism has been prominently used in crop pest recognition to enhance the performance of deep learning models. A parallel attention mechanism integrated with a deep network showed improvements in the accuracy \cite{zhao2022crop}. In another work, the in-field pest recognition performance is boosted by combining spatial and channel attention \cite{yang2021recognizing}. Attention mechanism embedded with lightweight networks obtained better classification accuracy in pest recognition domain \cite{chen2021crop}. 

In summary, a lightweight network-based pest recognition system is ideal for in-field utilization as it is fast and consumes less memory. However, it is noticed that these often failed to attain expected classification accuracy. Attention can fix the performance issue of any deep learning model  \cite{takalkar2021lgattnet}, which creates a lead for this work.
\vspace{-0.2cm}
\section{Proposed model}
\label{sec:proposedmethod}
This section presents the proposed double attention-based lightweight pest recognition network in detail.
\vspace{-0.2cm}
\subsection{The Lightweight Network}
Figure \ref{fig:overallarchitecture} illustrates an overview of the proposed lightweight network with the example activation maps for an example input. In the figure, the $A\_1_{before}$ and $A\_1_{after}$ are the activation maps obtained before and after the first double attention layer. Similarly, the activation maps before and after the second double attention layer are given in $A\_2_{before}$ and $A\_2_{after}$, respectively. The $A\_3_{final}$ is the activation map generated by the last convolutional layer of the proposed architecture.

\begin{table}[t]
	\centering
	\caption{Layer organization of the proposed lightweight pest recognition architecture.}
	\begin{tabular}{|l|l|l|l|l|} 
		\hline
		\textbf{\#} & \textbf{Input} & \textbf{Layer} & \textbf{Output} & \textbf{Stride} \\
		\hline
		1 & $224\times224\times3$ & conv $3\times3$ & $112\times112\times32$ & 2\\
		2 & $112\times112\times32$ & channel reduction & $112\times112\times16$ & 1\\
		3 & $112\times112\times16$ & down sampling & $56\times56\times24$ & 2\\
		4 & $56\times56\times24$ & $1\times$inverted residual & $56\times56\times24$ & 1\\
		5 & $56\times56\times24$ & down sampling & $28\times28\times32$ & 2\\
		6-7 & $28\times28\times32$ & $2\times$inverted residual & $28\times28\times32$ & 1\\
		8 & $28\times28\times32$ & down sampling & $14\times14\times64$ & 2\\
		9-10 & $14\times14\times64$ & $2\times$inverted residual & $14\times14\times64$ & 1\\
		11 & $14\times14\times64$ & double attention & $14\times14\times64$ & 1\\
		12 & $14\times14\times64$ & channel expansion & $14\times14\times96$ & 1\\
		13 & $14\times14\times96$ & $1\times$inverted residual & $14\times14\times96$ & 1\\
		14 & $14\times14\times96$ & double attention & $14\times14\times96$ & 1\\
		15 & $14\times14\times96$ & down sampling & $7\times7\times160$ & 2\\
		16-17 & $7\times7\times160$ & $2\times$inverted residual & $7\times7\times160$ & 1\\
		18 & $7\times7\times160$ & channel expansion & $7\times7\times320$ & 1\\
		19 & $7\times7\times320$ & conv $1\times1$ & $7\times7\times1280$ & 1\\
		20 & $7\times7\times1280$ & avgpool & $1\times1\times1280$ & -\\
		21 & $1\times1\times1280$ & dropout \& softmax & $1\times1\times{k}$ & -\\
		\hline 
	\end{tabular}
	\label{table:proposedapproach}
\end{table}

The comprehensive details of each layer in the proposed network, where the MobileNetv2 is used as the backbone, are individually illustrated in Table \ref{table:proposedapproach}. Altogether, this architecture has 21 primary layers: convolution, channel reduction, down sampling, inverted residual and channel expansion, average pooling, and the final dropout and softmax layers. Inspired by the key architectural concepts of the MobileNetv2 deep network, the inverted residual and depthwise separable convolution that are unique to MobileNetv2 are exploited in the proposed architecture. For example, inverted residual layers are built in a specific way which expands the channels six times and apply the depthwise separable convolution in order to learn the features efficiently. In addition, the inverted residual layers create the shortcut connection between the input and output same as ResNet\cite{he2016deep} to enable the residual learning. This allows to build deeper networks. In depthwise separable convolution, a depthwise convolution operation is first performed and followed by a pointwise convolution operation. Notably, as demonstrated in \cite{howard2017mobilenets}, depthwise separable convolution has the benefit of computational efficiency due to the fact that the depthwise convolution can potentially reduce the spatial dimensions.

The depthwise separable convolution in the higher channel dimension with a similar structure is used to build both channel expansion and downsampling layers. However, in channel expansion layers, a higher number of pointwise filters are defined to increase the output channels. In contrast, for downsampling layers, the stride is set to 2 along with a depthwise convolution that reduces the spatial dimensions by half while the subsequent pointwise filters increasing the channels. In addition, the first standard conv $3\times3$ layer with the stride set to 2 reduces the spatial dimension by half and the subsequent channel reduction layer (layer 2) reduces the channels by half. This drastic reductions in dimensions makes early layers efficient.

As given in Figure \ref{fig:overallarchitecture} and Table \ref{table:proposedapproach}, two double attention modules are systematically affixed as layer 11 and 14 in this architecture which is explained in the next subsection.  

\begin{figure*}[t]
	\begin{center}
		\includegraphics[width=\textwidth]{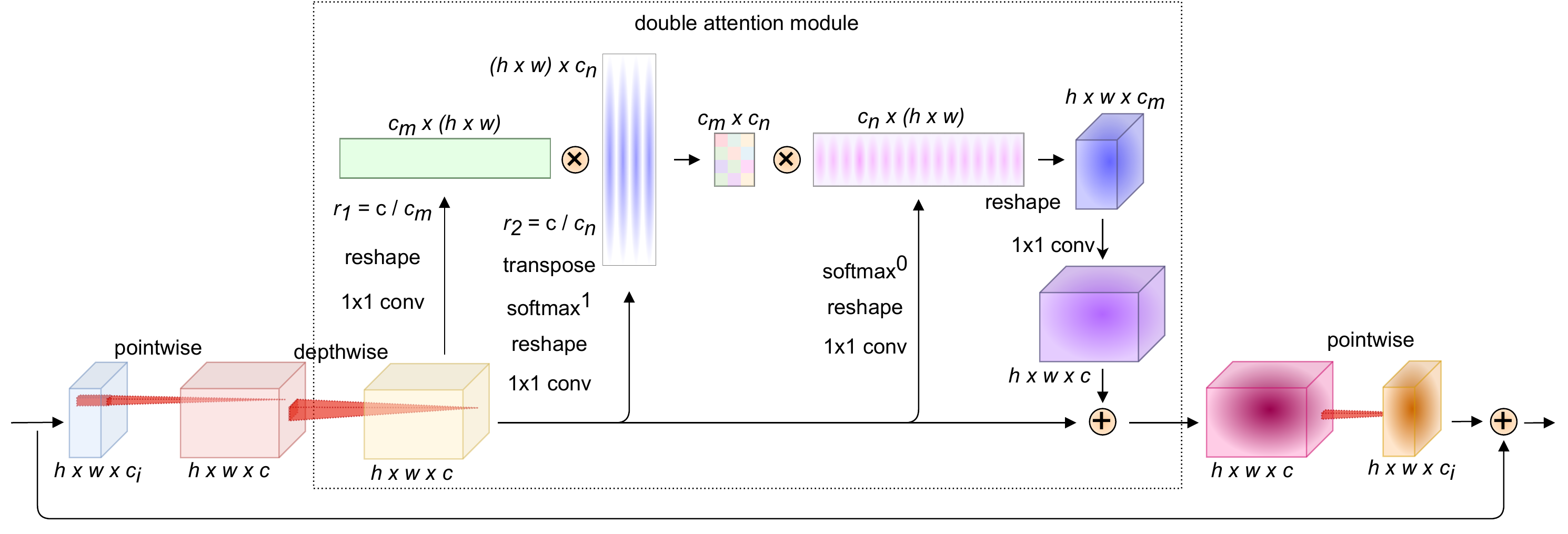}
	\end{center}
	
	\caption{The double attention layer used in our proposed architecture.}
	
	\label{fig:doubleattention}
\end{figure*}
\vspace{-0.2cm}
\subsection{Double Attention Layer}
Figure \ref{fig:doubleattention} shows the complete structure of the double attention layer proposed in \cite{chen20182}, which is known for capturing long-range relations. Generally, for any image classification tasks, it is confirmed that a lightweight network integrated with double attention modules obtained superior results over a much larger traditional deep network. For instance, a much smaller double attention enabled ResNet-50 model outperformed the ResNet-152 model on the ImageNet-1k dataset for a classification task. Here, the informative global features extracted from the whole image are manipulated in a two-phased attention scheme, namely aggregation and propagation. The double attention module can easily be plugged into any deep network, allowing to pass useful information learned from the entire feature space effectively to subsequent convolution layers. In the aggregation step, the key features are extracted from the entire space using a second-order attention pooling mechanism. Meanwhile, in the propagation step, a subset of the meaningful features corresponding to each spatial location is adaptively picked and dispersed using another attention mechanism.  

Two hyperparameters $r_1$ and $r_2$ can be observed in this mechanism, the first one is to reduce the channels from $c$ to $c_m$ and the second one to reduce the channels from $c$ to $c_n$. Both of these parameters are fine-tuned to obtain the best performing combination. However, similar to the procedure followed in the original paper, a common hyperparameter is used to tune both of these hyperparameters. The double attention layers are built deeply integrating the double attention module into the inverted residual layers as shown in Figure \ref{fig:doubleattention}. The double attention module is placed next to the depthwise convolution to exploit the higher dimensional feature space (six times the number of input channels). The importance of this placement is further discussed in the ablative analysis, in subsection \ref{sec:ablative}. More importantly, both of the double attention modules are placed in deeper layers since it is confirmed in the original paper that the double attention modules perform well when they are attached to deeper layers.
\vspace{-0.2cm}
\section{Experiments}
\label{sec:experiments}
\vspace{-0.2cm}
\subsection{Datasets}
Two publicly available pest datasets containing images with natural backgrounds, published in \cite{li2020crop} and \cite{cheng2017pest}, are used. Throughout the paper, the former (\cite{li2020crop}) is referred to as the D1 dataset while the latter (\cite{cheng2017pest}) is mentioned as the D2 dataset.

\textbf{D1 dataset:} The D1 dataset consists of the data for the ten most prevalent crop pests with fast reproductive rates and causing significant yield losses. The images are collected using internet search through the most popular search engines, such as Google, Baidu, Yahoo, and Bing. The authors of the D1 dataset also use mobile phone cameras (i.e., Apple 7 Plus) to collect additional pest images from the environment. The published D1 dataset contains 5,869 images for ten pests, namely Cydia Pomonella (415), Gryllotalpa (508), Leafhopper (426), Locust (737), Oriental Fruit Fly (468), Pieris Rapae Linnaeus (566), Snail (1072), Spodoptera litura (437), Stink Bug (680) and Weevil (560). 

\textbf{D2 dataset:} Cheng et al. \cite{cheng2017pest} constructed the D2 pest dataset by picking ten species of pests from the original dataset used in \cite{xie2015automatic}. The original dataset contains high-resolution images in the size of $1280\times960$, captured using colour digital cameras (i.e., Canon and Nikon). However, in the D2 dataset, the resolution of the original images is reduced to $256\times256$. There are 545 image samples of ten species, namely Elia Sibirica (55), Cifuna Locuples (55), Cletus Punctiger (55), Cnaphalocrocis Medinalis (54), Colaphellus Bowvingi (51), Dolerus Tritici (55), Pentfaleus Major (55), Pieris Rapae (55), Sympiezomias Velatus (55) and Tettigella Viridis (55), are included in this dataset.

\textbf{D1\textsubscript{500} dataset:} To evaluate the proposed network with small data, a subset of D1 dataset with 500 image samples is constructed by  randomly selecting 50 samples per class. 
\vspace{-0.2cm}
\subsection{Implementation and Training}
The PyTorch deep learning framework and Google Colab\footnote{https://colab.research.google.com/} platform are utilized to implement the proposed pest recognition model and other existing state-of-the-art lightweight models used for comparison. The pre-trained ImageNet weights available with the PyTorch framework are loaded before fine-tuning all the state-of-the-art networks on pest datasets. Similarly, the proposed model is trained on the ImageNet dataset before the fine-tuning process. 

During the training of all the models on pest datasets, the number of epochs, batch size, imput image size are set to 50, 8 and $224\times224$ with centre cropped, respectively. The stochastic gradient descent (SGD) optimizer with an initial learning rate of 0.001 is applied for all the models, except for MNasNet (0.005) and ShuffleNet (0.01) when they are trained with dataset D2 and D1\textsubscript{500}. The momentum and weight decay of the SGD are set to 0.9 and $1e\textsuperscript{-4}$, respectively.
\vspace{-0.2cm}
\subsection{Results}
In this section, the evaluation results obtained for the proposed double attention-based lightweight pest recognition architecture under diverse environments are discussed. The results are reported using two evaluation metrics: average accuracy and F1-measure. All the experimental results reported in this paper used 5-fold cross-validation procedure.
\vspace{-0.2cm}
\begin{table*}[!h]
	\centering
	
	\caption{Comparison of the average accuracy (Acc) and F1-measure (F1) obtained for the proposed model and other state-of-the-art lightweight models on D1, D2 and D1\textsubscript{500} pest datasets. The number of parameters (Params [Millions]) and floating point operations (FLOPs [GMAC]) are also given.}
	
	\begin{tabular}{|l|c|c|c|c|c|c|c|c|}
		\hline
		\multicolumn{1}{|c|}{\multirow{2}{*}{Method}} & \multicolumn{1}{c|}{\multirow{1}{*}{Params}} & \multicolumn{1}{c|}{\multirow{1}{*}{FLOPs}} & 
		\multicolumn{2}{c|}{D1[5869]} & \multicolumn{2}{c|}{D2[545]} & \multicolumn{2}{c|}{D1\textsubscript{500}[500]}\\
		\cline{4-9}
		
		&[Millions]&[GMAC]&F1&Acc&F1&Acc&F1&Acc\\
		\hline
		ShuffleNetv2 & 1.26 & 0.15 & 93.47 & 93.90 & 97.94 & 97.94 & 85.60 & 85.60 \\
		SqueezeNet & 0.74 & 0.74 & 91.40 & 91.85 & 95.19 & 95.22 & 79.11 & 78.50 \\
		MobileNetv2 & 2.24 & 0.32 & 95.32 & 95.69 & 98.72 & 98.71 & 89.74 & 89.80 \\
		MNasNet & 3.12 & 0.33 & 95.80 & 96.16 & 94.23 & 94.30 & 87.22 & 87.40 \\
		MobileNetv3\_{Large} & 4.21 & 0.23 & 96.04 & 96.34 & 97.60 & 97.61 & 89.02 & 89.20\\
		MobileNetv3\_Small & 1.53 & 0.06 & 94.50 & 94.58 & 95.81 & 95.77 & 85.40 & 85.60 \\
		\textbf{Proposed Method}&2.56&0.38&\textbf{96.37}&\textbf{96.61}&\textbf{99.09}&\textbf{99.08}&\textbf{91.55}&\textbf{91.60}\\
		\hline
	\end{tabular}
	
	\label{table:results}
\end{table*}

Table \ref{table:results} presents the comparison results obtained on D1, D2 and D1\textsubscript{500} datasets. In addition to reporting average accuracy and F1-measure, the number of parameters in millions and floating-point operations (FLOPs) of each model are also provided. The proposed double attention-based pest recognition approach achieved the average accuracy of 96.61\%, 99.09\% and 91.55\% on D1, D2 and D1\textsubscript{500} datasets, respectively. Compared to other lightweight models, the proposed approach showed a comprehensive improvement. When it comes to small training data, the proposed model outperformed the compared models, at least by 0.37\% on D2 and 1.8\% on D1\textsubscript{500}. This indicates that the proposed method can produce high plant pests classification capability under data constraint conditions. The proposed model achieved very high F1 measures on all datasets, especially 96.61\% on the D1 dataset, which reveals the feasibility of this method in dealing with data with imbalanced class samples. While the MobileNetv3\_Large model achieved the second-highest average accuracy on the D1 dataset, the MobileNetv2 outperformed it on the other two datasets. This exhibits the superiority of the MobileNetv2 model on small datasets in comparison to other state-of-the-art lightweight networks.

\begin{figure*}[!h]
	\begin{center}
		\includegraphics[width=\textwidth]{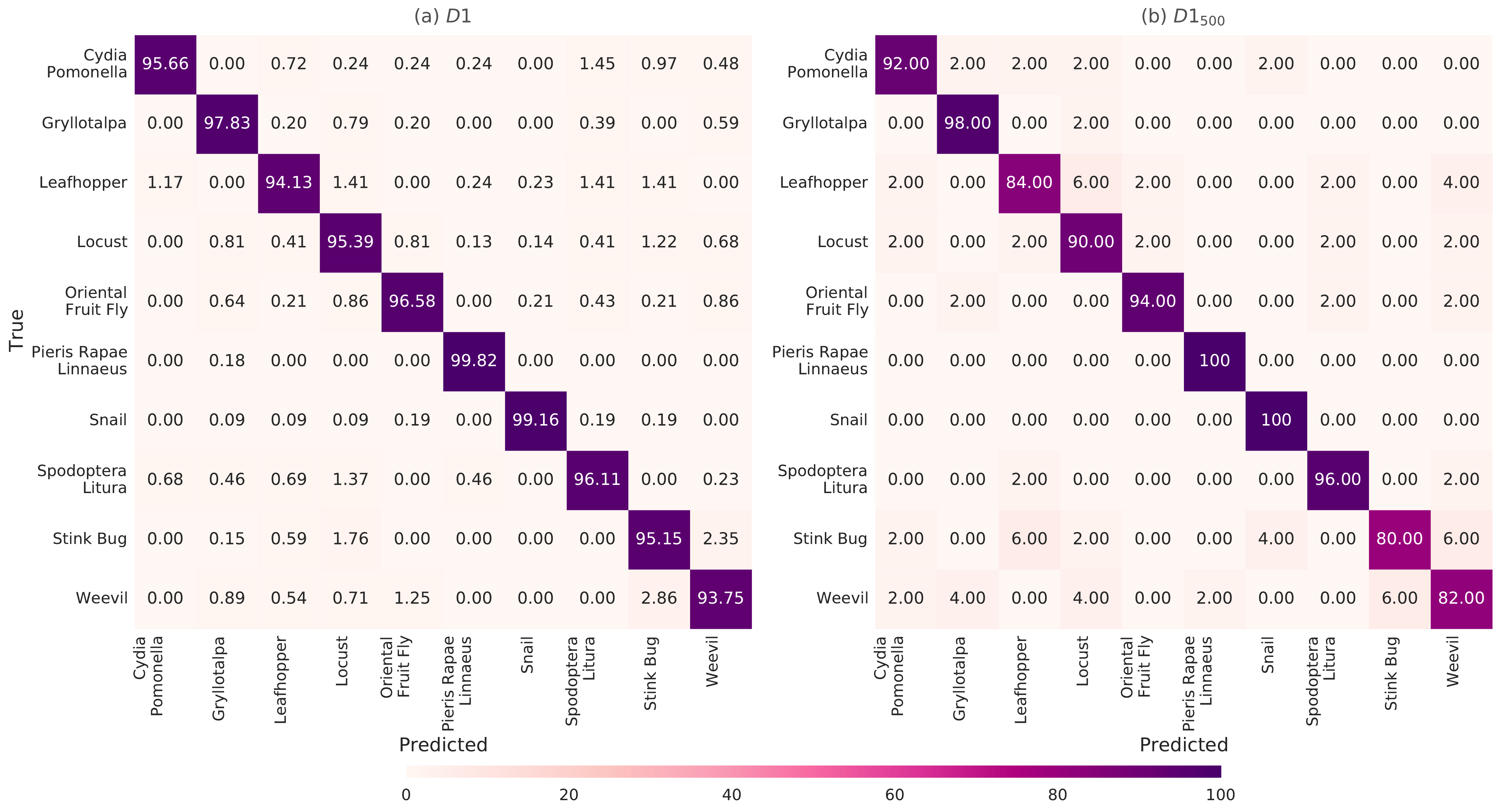}
	\end{center}
	
	\caption{Confusion matrix obtained for the proposed model on D1 and D1\textsubscript{500} pest datasets.}
	
	\label{fig:confusionmatrix}
\end{figure*}

\begin{figure*}[!h]
	\begin{center}
		\includegraphics[width=0.9\textwidth]{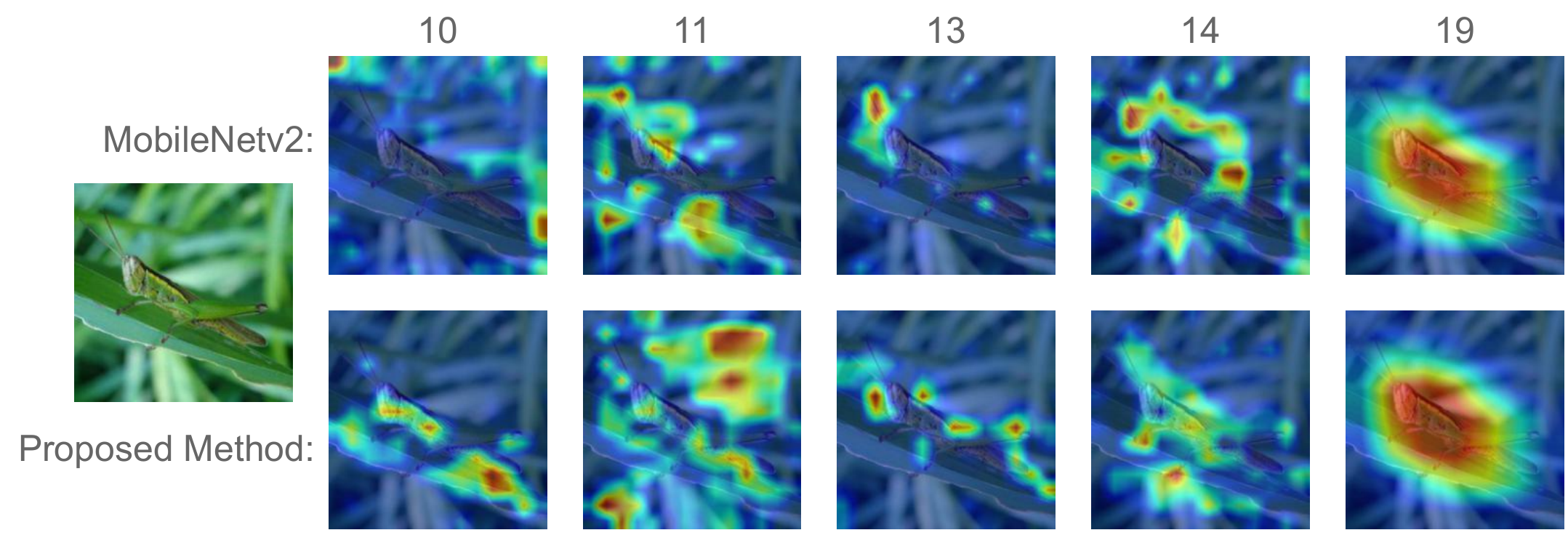}
	\end{center}
	
	\caption{Comparison of the activation maps generated by the proposed model and MovileNetv2 for an example pest image taken from D1 dataset.}
	
	\label{fig:activations}
\end{figure*}

The number of parameters for the proposed method is 2.56 million, which is slightly more than the original MobileNetv2 model's 2.24 million parameters.  However, the proposed model holds 0.56 and 1.65 million parameters lesser than MNasNet and MobileNetv3\_large, respectively, which is a significant improvement. The FLOPs counts to run a single instance of all the compared models are also compared, and one can see that the proposed model manifests comparable number of FlOPs. The models with 0.4 GMAC or less are generally considered as the mobile-size baseline [15]. Hence, it can be concluded that the proposed pest recognition model is adaptive for mobile deployment and comprehensively used in in-field settings.

In order to further emphasize the class-wise performance of the proposed model on D1 and D1\textsubscript{500} datasets, the confusion matrices are presented in Figure \ref{fig:confusionmatrix}. Overall, the proposed pest recognition model showed high class-wise accuracies for most of the pest classes. Compared to the other pest classes, on both datasets, the Pieris Rapae Linnaeus, Snail and Gryllotalpa pest classes consistently obtained high average accuracies because of their unique characteristics. The low average accuracies are recorded for the pest classes, such as Weevil, Leafhopper and Stink bug. The accuracies are greatly deprived for these pest classes when reducing the number of samples in the dataset (i.e., 5869 samples of D1 to 500 samples of D1\textsubscript{500}). For example, the reduction of 15.15\% for Stink bug and 11.75\% for Weevil are witnessed from D1 dataset to D1\textsubscript{500} dataset. Further, the majority of the misclassified samples from Stink bug and Weevil pest classes are confused between those two classes due to a substantial amount of shared characteristics. For instance, 6.00\% of the Stink bug pests are confused with the Weevil and vice versa.

Figure \ref{fig:activations} provides the qualitative results of the proposed pest recognition model, where the activation maps obtained in different layers are compared. The activation maps are generated for a Locust pest sample, which is correctly classified by both the proposed method and MobileNetv2. The activation maps for the learned feature representations by the proposed model and MobileNetv2 generated on layers 10, 11, 13, 14 and 19 are compared. As can be seen, the activations obtained by the proposed model convey better meanings in respective layers, compared to MobileNetv2. Fro example, it can be observed that the pest parts are more accurately captured in the activation maps generated by the attention layers of the proposed method (layers 11 and 14). This behaviour positively contributes to the overall pest recognition performance.

\begin{table}[t]
	\centering
	
	\caption{The results of an ablative study conducted to select the locations of the double attention modules and the values of the hyperparameters $r_1$ and $r_2$. The average accuracies alongside the number of parameters and FLOPs are presented.}
	
	\begin{tabular}{|c|c|c|c|c|c|c|}
		\hline
		\multicolumn{2}{|c|}{\multirow{1}{*}{Locations}} & 
		\multicolumn{1}{c|}{\multirow{2}{*}{$r_1, r_2$}} & 
		\multicolumn{1}{c|}{\multirow{1}{*}{Params}} & 
		\multicolumn{1}{c|}{\multirow{1}{*}{FLOPs}} & 
		\multicolumn{1}{c|}{\multirow{2}{*}{Acc}}\\
		\cline{1-2}
		\multicolumn{1}{|c|}{\multirow{1}{*}{$l_1$}}&
		\multicolumn{1}{c|}{\multirow{1}{*}{$l_2$}}&
		&(Millions)&(GMAC)&\\
		\hline
		10&11&4&3.80&0.3783&$73.03\pm0.25$\\
		11&14&4&3.99&0.4145&$73.16\pm0.38$\\
		13&14&4&4.17&0.4507&$72.93\pm0.26$\\
		10&11&6&3.70&0.3590&$72.78\pm0.23$\\
		11&14&6&3.83&0.3831&$\textbf{73.22}\pm0.35$\\
		13&14&6&3.95&0.4073&$73.03\pm0.36$\\
		10&11&8&3.65&0.3494&$72.74\pm0.38$\\
		11&14&8&3.75&0.3675&$73.00\pm0.24$\\
		13&14&8&3.84&0.3856&$73.03\pm0.48$\\
		\hline
		11\textsuperscript{*}&14\textsuperscript{*}&1&3.56&0.3308&$73.08\pm0.29$\\
		\hline
	\end{tabular}
	
	\label{table:ablation}
\end{table}

\subsection{Ablative Study}
\label{sec:ablative}
In this subsection, an ablative analysis is performed to explain the process behind determining the ideal location of the double attention modules in the proposed lightweight network. Table \ref{table:ablation} gives the recorded average accuracies for various combinations of double attention location and $r_1$ and $r_2$ hyperparameter values. The selection of the positions of the double attention modules and the fine-tuning of the hyperparameters $r_1$ and $r_2$ are performed on the more sophisticated CIFAR100 dataset. The locations of the double attention modules are fixed in deeper layers as it is empirically endorsed. However, it is possible to integrate two subsequent double attention layers between 9 and 11 (set of units processing 64 channels) or 12 and 14 (processing 96 channels) or one each in those sets. Although integrating into the second set would process more channels, applying attention interleaved can help in learning the attention well. Hence, Multiple combinations of the double attention module locations (10,11), (11,14) and (13,14) and the hyperparameters $r_1$ and $r_2$ values of 4, 6 and 8 are defined to grid search the best performing tuple with the highest average accuracy. In addition, the number of parameters and FLOPs are also recorded. The best model accuracy is achieved when the double attention modules are affixed in locations 11 and 14 along with six as the value for hyperparameters $r_1$ and $r_2$, which are then utilized to build the proposed model.

The last row of Table \ref{table:ablation} shows the accuracy obtained when the double attention modules are integrated directly to the MobileNetv2-backbone after 11\textsuperscript{th} and 14\textsuperscript{th} layers, instead of deeply integrating into the inverted residual layers. As can be seen, the proposed method achieves better accuracy of 73.22\% over this, which achieves only 73.08\% accuracy. The hyperparameters $r_1$ and $r_2$ were set to 1 in this case, to maintain the same feature dimensions within the attention module. This shows the attention learning is more effective in the higher dimensional feature space and importance of integrating the double attention module deeply into the inverted residual layer as shown in Figure \ref{fig:doubleattention}. 

\section{Conclusion}
\label{sec:conclusion}
\vspace{-0.1cm}
In order to enable in-field pest recognition model trained with small data, in this paper, a novel double attention-based lightweight architecture is presented. As the MobileNetv2 is utilised as the backbone, this lightweight network showed useful characteristics of less memory consumption and faster inference. The proposed architecture also exploited the benefit of increased performance derived by deploying a double attention mechanism. The evaluation results on two benchmark datasets and one reduced dataset proves the the promising capability of the proposed approach in effectively classify plant pests in in-field settings with small data for training. Specifically, the recognition accuracies on small data trends to prevail, by achieving 99.08\% and 91.60\%. The comparative analysis demonstrates that the proposed approach is superior over existing pest recognition methods. In the future, as new pest classes appear down the line, continual learning ability can be infused to further extend the in-field applicability of the proposed approach. The proposed framework can effectively spot smaller objects and objects confusing with their background. It mainly enhances the features in the identified areas for better performance. This trait is helpful for other applications like chip defect classification, where the defects are tiny and often subtle. In addition, the lightweight aspect of this network enable edge devise deployment.


\end{document}